\documentclass[11pt,a4paper]{article}
\usepackage[hyperref]{udst2018}
\usepackage{times}
\usepackage{latexsym}
\usepackage{amsmath}
\usepackage{amsfonts}

\usepackage{url}

\usepackage{makecell}

\aclfinalcopy 
\def\aclpaperid{***} 

\title{Towards JointUD: Part-of-speech Tagging and Lemmatization using Recurrent Neural Networks}

\author{Gor Arakelyan, Karen Hambardzumyan, Hrant Khachatrian \\
  YerevaNN / 9, Charents str., apt. 38, Yerevan, Armenia \\
  Yerevan State University / 1, Alex Manoogian str., Yerevan, Armenia \\
  {\tt \{gor.arakelyan,mahnerak,hrant\}@yerevann.com}  \\}

\date{}
\hypersetup{draft}
\begin{document}
\maketitle

\newcommand{\udst}[0]{\emph{CoNLL 2018 UD Shared Task}}

\begin{abstract}
This paper describes our submission to \udst{}. We have extended an LSTM-based neural network designed for sequence tagging to additionally generate character-level sequences. The network was jointly trained to produce lemmas, part-of-speech tags and morphological features. Sentence segmentation, tokenization and dependency parsing were handled by UDPipe 1.2 baseline. The results demonstrate the viability of the proposed multitask architecture, although its performance still remains far from state-of-the-art.

\end{abstract}

\section{Introduction}
The Universal Dependencies project \cite{ud} aims to collect consistently annotated treebanks for many languages. Its current version (2.2) \cite{ud22data} includes publicly available treebanks for 71 languages in CoNLL-U format. The treebanks contain lemmas, part-of-speech tags, morphological features and dependency relations for every word.

Neural networks have been successfully applied to most of these tasks and produced state-of-the-art results for part-of-speech tagging and dependency parsing. Part-of-speech tagging is usually defined as a sequence tagging problem and is solved with recurrent or convolutional neural networks using word-level softmax outputs or conditional random fields \cite{LSTMCRFNER, ERDilatedConv, NERBiLSTMCNN}. \citet{Reimers:2017:EMNLP} have studied these architectures in depth and demonstrated the effect of  network hyperparameters and even random seeds on the performance of the networks.

Neural networks have been applied to dependency parsing since 2014 \cite{ChenManning}. The state-of-the-art in dependency parsing is a network with deep biaffine attention module, which won \emph{CoNLL 2017 UD Shared Task} \cite{StanfordCoNLL2017}. 

\citet{jPTDP1} used a neural network to jointly learn POS tagging and dependency parsing. To the best of our knowledge, lemma generation and POS tagging have never been trained jointly using a single multitask architecture.

This paper describes our submission to \udst{}. We have designed a neural network that jointly learns to predict part-of-speech tags, morphological features and lemmas for the given sequence of words. This is the first step towards \textit{JointUD}, a multitask neural network that will learn to output all labels included in UD treebanks given a tokenized text. Our system used UDPipe 1.2 \cite{udpipe} for sentence segmentation, tokenization and dependency parsing.

Our main contribution is the extension of a sequence tagging network by \citet{Reimers:2017:EMNLP} to support character-level sequence outputs for lemma generation. The proposed architecture was validated on nine UD v2.2 treebanks. The results are generally not better than the UDPipe baseline, but we did not extensively tune the network to squeeze most out of it. Hyperparameter search and improved network design are left for the future work.

\section{System Architecture}

Our system used in \udst{} consists of two parts. First, it takes the raw input and produces CoNLL-U file using UDPipe 1.2. Then, if the corresponding neural model exists, the columns corresponding to lemma, part-of-speech and morphological features are replaced by the predictions of the neural model. Note that UDPipe 1.2 did not use the POS tags and lemmas produced by our neural model. We did not train neural models for all treebanks, so most of our submissions are just the output of UDPipe. 

The codename of our system in the Shared Task was \textit{ArmParser}. The code is available on GitHub\footnote{ \url{https://github.com/YerevaNN/JointUD/}}.

\section{Neural model}

In this section we describe the neural architecture that takes a sequence of words and outputs lemmas, part-of-speech tags, and 21 morphological features. POS tag and morphological feature prediction is done using a sequence tagging network from \cite{Reimers:2017:EMNLP}. To generate lemmas, we extend the network with multiple decoders similar to the ones used in sequence-to-sequence architectures.

Suppose the sentence is given as a sequence of words $w_1, \ldots, w_n$. Each word consists of characters $w_i = c_i^1 \dots c_i^{n_i}$. For each $w_i$, we are given its lemma as a sequence of characters: $l_i = l_i^1 \ldots l_i^{m_i}$, POS tag $p_i \in P$, and 21 features $f_i^1 \in F^1, \ldots, f_i^{21} \in F^{21}$. The sets $P, F^1, \ldots, F^{21}$ contain the possible values for POS tags and morphological features and are language-dependent: the sets are constructed based on the training data of each language. Table \ref{table:ewt} shows the possible values for POS tags and morphological features for \textit{English - EWT} treebank.

\begin{table*}[t]
\begin{tabular}{l|l}
Tag    & Values                                    \\\hline
POS & \makecell[l]{PROPN (6.328\%), PUNCT (11.574\%), ADJ (6.102\%), NOUN (16.997\%), \\ VERB (11.254\%), DET (7.961\%), ADP (8.614\%), AUX (6.052\%), \\ PRON (9.081\%), PART (2.721\%), SCONJ (1.881\%), NUM (1.954\%), ADV (5.158\%), \\ CCONJ (3.279\%), X (0.414\%), INTJ (0.336\%), SYM (0.295\%)}   \\\hline
Number & Sing (27.357\%), Plur (6.16\%), None (66.483\%)  \\\hline
Degree & Pos (5.861\%), Cmp (0.308\%), Sup (0.226\%), None (93.605\%)  \\\hline
Mood & Ind (7.5\%), Imp (0.588\%), None (91.912\%)  \\\hline
Tense & Past (4.575\%), Pres (5.316\%), None (90.109\%)  \\\hline
VerbForm & Fin (9.698\%), Inf (4.042\%), Ger (1.173\%), Part (2.391\%), None (82.696\%)  \\\hline
Definite & Def (4.43\%), Ind (2.07\%), None (93.5\%)  \\\hline
Case & Acc (1.284\%), Nom (4.62\%), None (94.096\%)  \\\hline
Person & 1 (3.255\%), 3 (5.691\%), 2 (1.396\%), None (89.658\%)  \\\hline
PronType & Art (6.5\%), Dem (1.258\%), Prs (7.394\%), Rel (0.569\%), Int (0.684\%), None (83.595\%)  \\\hline
NumType & Card (1.954\%), Ord (0.095\%), Mult (0.033\%), None (97.918\%)  \\\hline
Voice & Pass (0.589\%), None (99.411\%)  \\\hline
Gender & Masc (0.743\%), Neut (0.988\%), Fem (0.24\%), None (98.029\%)  \\\hline
Poss & Yes (1.48\%), None (98.52\%)  \\\hline
Reflex & Yes (0.049\%), None (99.951\%)  \\\hline
Foreign & Yes (0.009\%), None (99.991\%)  \\\hline
Abbr & Yes (0.04\%), None (99.96\%)  \\\hline
Typo & Yes (0.052\%), None (99.948\%)
\end{tabular}
\caption{The values for part-of-speech and morphological features for \textit{English - EWT} treebank.}
\label{table:ewt}
\end{table*}

The network consists of three parts: embedding layers, feature extraction layers and output layers.

\subsection{Embedding layers}
\label{sec:embedding}
By $\textrm{Emb}^d(a)$ we denote a $d$-dimensional embedding of the integer $a$. Usually, $a$ is an index of a word in a dictionary or an index of a character in an alphabet.

Each word $w_i$ is represented by a concatenation of three vectors: $e(w_i) = \left(e_{word}(w_i), e_{casing}(e), e_{char}(w) \right)$. The first vector, $e_{word}(w_i)$ is a 300-dimensional pretrained word vector. In our experiments we used FastText vectors \cite{fasttext} released by Facebook\footnote{\url{https://github.com/facebookresearch/fastText/blob/master/pretrained-vectors.md}}. The second vector, $e_{casing}(w_i)$, is a one-hot representation of eight casing features, described in Table \ref{table:casing}.

The third vector, $e_{char}(w_i)$ is a character-level representation of the word. We map each character to a randomly initialized 30-dimensional vector $\widehat{c}_i^j = Emb^{30}(c_i^j)$, and apply a bi-directional LSTM on these embeddings. $e_{char}(w_i)$ is the concatenation of the 25-dimensional final states of two LSTMs. 

The resulting $e(w_i)$ is a 358-dimensional vector.

\begin{table*}[t]
\centering
\begin{tabular}{l|l}
\textit{numeric} & All characters are numeric \\
\textit{mainly numeric} & More than 50\% of characters are numeric \\
\textit{all lower} & All characters are lower cased \\
\textit{all upper} & All characters are upper cased \\
\textit{initial upper} & The first character is upper cased \\
\textit{contains digit} & At least one of the characters is digit \\
\textit{other} & None of the above rules applies \\
\textit{padding} & This is used for the padding placeholders for short sequences 
\end{tabular}
\caption{Casing features used in the embedding layer.}
\label{table:casing}
\end{table*}

\subsection{Feature extraction layers}

We denote a recurrent layer with inputs $x_1, \ldots, x_n$ and hidden states $h_1, \ldots, h_n$ by $h_i = RNN(x_i, h_{i-1})$. We use two types of recurrent cells: LSTM \cite{lstm} and GRU \cite{gru}.

We apply three layers of LSTM with 150-dimensional hidden states on the embedding vectors:
\begin{align*}
&h^j_i = LSTM\left(h^{j-1}_i, h^{j-1}_{i-1}\right) & j=1,2,3
\end{align*}
where $h^0_i = e(w_i)$. We also apply 50\% dropout before each LSTM layer. 

The obtained 150-dimensional vectors represent the words with their contexts, and are expected to contain necessary information about the lemma, POS tag and morphological features.

\subsection{Output layers}
\subsubsection{POS tags and features}
Part-of-speech tagging and morphological feature prediction are word-level classification tasks. For each of these tasks we apply a linear layer with softmax activation.
\begin{align*}
& \widetilde{p_i} = softmax(W_p h^3_i + b_p) \\
& \widetilde{f_i^k} = softmax\left(W_{f^k} h^3_i + b_{f^k}\right) & k=1,\ldots,21
\end{align*}

The dimensions of the matrices $W_p$, $W_{f^k}$ and vectors $b_p$, $b_{f^k}$ depend on the training set for the given language: $W_p \in \mathbb{R}^{|P| \times 150}$, $W_{f^k} \in \mathbb{R}^{|F^k| \times 150}$, $k=1,\ldots,21$.
So we end up with 22 cross-entropy loss functions: 
\begin{align*}
& L_p = \frac{1}{n}\sum_{i=1}^n{ce(\widetilde{p_i}, p_i)} \\
& L_{f^k} = \frac{1}{n}\sum_{i=1}^n{ce\left(\widetilde{f_i^k}, f_i^k\right)} & k=1,\ldots,21
\end{align*}

\subsubsection{Lemma generation}
This subsection describes our main contribution. In order to generate the lemmas for all words, we add one GRU-based decoder per each word. These decoders share the weights and work in parallel. The $i$-th decoder outputs $\widetilde{l}_i^1, \ldots, \widetilde{l}_i^{m_i}$, the predicted characters of the lemma of the $i$-th word. We denote the inputs to the $i$-th decoder by $x_i^1, \ldots, x_i^{m_i}$. Each of $x_i^j$ is a concatenation of four vectors: $x_i^j = \left(h_i^3, \widehat{c}_i^j, \pi_i^j, \widehat{l}_i^{j-1} \right)$.
\begin{enumerate}
\item $h_i^3$ is the representation of the $i$-th word after feature extractor LSTMs. This is the only part of $x_i^j$ vector that does not depend on $j$. This trick is important to make sure that word-level information is always available in the decoder. 
\item $\widehat{c}_i^j = Emb^{30}(c_i^j)$ is the same embedding of the $j$-th character of the word used in the character-level BiLSTM described in Section \ref{sec:embedding}.
\item $\pi_i^j$ is some form of positional encoding. It indicates the number of characters remaining till the end of the input word: $\pi_i^j = Emb^5(n_i - j + 1)$. Positional encodings were introduced in \cite{end2end-memory-networks} and were successfully applied in neural machine translation \cite{convs2s:fairseq2seq,attention-is-all-you-need}.
\item $\widehat{l}_i^{j-1}$ is the indicator of the previous character of the lemma. During training it is the one-hot vector of the ground-truth: $\widehat{l}_i^{j-1} = onehot(l_i^{j-1})$. During inference it is the output of the GRU in the previous timestep $\widehat{l}_i^{j-1} = \widetilde{l}_i^{j-1}$.
\end{enumerate}

These inputs are passed to a single layer of GRU network. The output of the decoder is formed by applying another dense layer on the GRU state:
\begin{align*}
s_i^j &= GRU(x_i^j, s_i^{j-1}) \\
\widetilde{l}_i^j &= W_o s_i^j + b_o
\end{align*}
Here, $s_i^j \in \mathbb{R}^{150}$, $W_o \in \mathbb{R}^{|C| \times 150}$, where $|C|$ is the number of characters in the alphabet. The initial state of the GRU is the output of the feature extractor LSTM: $s_i^0 = h_i^3$. All GRUs share the weights.

The loss function for lemma output is:
\begin{align*}
L_l = \frac{1}{n} \sum_{i=1}^{n}{\frac{1}{n_i} \sum_{j=1}^{n_i}{ce\left(\widetilde{l}_i^j, l_i^j \right)}}
\end{align*}

\subsection{Multitask loss function}
The combined loss function is a weighted average of the loss functions described above:
\begin{align}
L = \lambda_l L_l + \lambda_p L_p + \sum_{k=1}^{21}{\lambda_{f^k} L_{f^k}}
\label{formula:loss}
\end{align}

The final version of our system used $\lambda_p = 0.2$ and $\lambda_l = \lambda_{f^k} = 1$ for every $k$.

\section{Experiments}
We have implemented the architecture defined in the previous section using Keras framework. Our implementation is based on the codebase for \cite{Reimers:2017:EMNLP}\footnote{\url{https://github.com/UKPLab/emnlp2017-bilstm-cnn-crf}}. The new part of the architecture (lemma generation) is quite slow. The overall training speed is decreased by more than three times when it is enabled. We have left speed improvements for future work.

To train the model we used RMSProp optimizer  with early stopping. The initial learning rate was $0.001$, and it was decreased to $0.0005$ since the seventh epoch. The training was stopped when the loss function was not improved on the development set for five consecutive epochs.

Due to time constraints, we have trained our neural architecture on just nine treebanks. These include three English and two French treebanks. 

Our system was evaluated on Ubuntu virtual machines in TIRA platform \cite{tira} and on our local machines using the test sets available on UD GitHub repository \cite{ud22testdata}.

The version we ran on TIRA had a bug in the preprocessing pipeline and was doubling new line symbols in the input text. Raw texts in UD v2.2 occasionally contain new line symbols inside the sentences. These symbols were duplicated due to the bug, and the sentence segmentation part of UDPipe treated them as two different sentences. The evaluation scripts used in \udst{} obviously penalized these errors. After the deadline of the Shared Task, we ran the same models (without retraining) on the test sets on our local machines without new line symbols.

Additionally, we locally trained models for two more non-Indo-European treebanks: Arabic PADT and Korean GSD. 

\subsection{Results}
Table \ref{table:las} shows the main metrics of \udst{} on the nine treebanks that we used for training our models. For each of the metrics we report five scores, two scores on our local machine (our model and UDPipe 1.2), and three scores from the official leaderboard\footnote{\url{
http://universaldependencies.org/conll18/results.html}} (our model, UDPipe baseline, the best score for that particular treebank). LAS metric evaluates sentence segmentation, tokenization and dependency parsing, so the numbers for our models should be identical to UDPipe 1.2. MLAS metric additionally takes into account POS tags and morphological features, but not the lemmas. BLEX metric evaluates dependency parsing and lemmatization. The full description of these metrics are available in \cite{udst:overview} and in \udst{}  website\footnote{\url{http://universaldependencies.org/conll18/evaluation.html}}. Table \ref{table:pos} compares the same models using another set of metrics that measure the performance of POS tagging, morphological feature extraction and lemmatization.


\begin{table*}[t]
\setlength{\tabcolsep}{0.3em} 
\scriptsize

\begin{tabular}{|l|c|c|c|c|c|c|c|c|c|c|c|c|c|c|c|}
\hline
\multicolumn{1}{|r|}{Metric}      & \multicolumn{5}{c|}{LAS}                                                                                                                      & \multicolumn{5}{c|}{MLAS}                                                                                                                     & \multicolumn{5}{c|}{BLEX}                                                                                                                     \\
\multicolumn{1}{|r|}{Environment} & \multicolumn{2}{c|}{Local}                             & \multicolumn{3}{c|}{TIRA}                                                            & \multicolumn{2}{c|}{Local}                             & \multicolumn{3}{c|}{TIRA}                                                            & \multicolumn{2}{c|}{Local}                             & \multicolumn{3}{c|}{TIRA}                                                            \\
\multicolumn{1}{|r|}{Model}       & \multicolumn{1}{c|}{Our} & \multicolumn{1}{c|}{UDPipe} & \multicolumn{1}{c|}{Our} & \multicolumn{1}{c|}{UDPipe} & \multicolumn{1}{c|}{Winner} & \multicolumn{1}{c|}{Our} & \multicolumn{1}{c|}{UDPipe} & \multicolumn{1}{c|}{Our} & \multicolumn{1}{c|}{UDPipe} & \multicolumn{1}{c|}{Winner} & \multicolumn{1}{c|}{Our} & \multicolumn{1}{c|}{UDPipe} & \multicolumn{1}{c|}{Our} & \multicolumn{1}{c|}{UDPipe} & \multicolumn{1}{c|}{Winner} \\\hline
English EWT                       & 77.12                    & 77.12                       & 65.69                    & 77.56                       & 84.57                       & 62.12                    & 68.27                       & 57.73                    & 68.70                       & 76.33                       & 66.35                    & 70.53                       & 60.73                    & 71.02                       & 78.44                       \\
English GUM                       & 74.21                    & 74.21                       & 60.89                    & 74.20                       & 85.05                       & 56.43                    & 62.66                       & 44.73                    & 62.66                       & 73.24                       & 58.75                    & 62.14                       & 48.54                    & 62.14                       & 73.57                       \\
English LinES                     & 73.08                    & 73.08                       & 60.52                    & 73.10                       & 81.97                       & 55.25                    & 64.00                       & 44.89                    & 64.03                       & 72.25                       & 57.91                    & 65.39                       & 47.24                    & 65.42                       & 75.29                       \\
French Spoken                     & 65.56                    & 65.56                       & 58.94                    & 65.56                       & 75.78                       & 51.50                    & 53.46                       & 47.08                    & 53.46                       & 64.67                       & 50.07                    & 54.67                       & 48.77                    & 54.67                       & 65.63                       \\
French Sequoia                    & 81.12                    & 81.12                       & 66.14                    & 81.12                       & 89.89                       & 64.56                    & 71.34                       & 55.68                    & 71.34                       & 82.55                       & 62.50                    & 74.41                       & 58.99                    & 74.41                       & 84.67                       \\
Finnish TDT                       & 76.45                    & 76.45                       & 58.65                    & 76.45                       & 88.73                       & 62.52                    & 68.58                       & 48.24                    & 68.58                       & 80.84                       & 38.56                    & 62.19                       & 28.87                    & 62.19                       & 81.24                       \\
Finnish FTB                       & 75.64                    & 75.64                       & 65.48                    & 75.64                       & 88.53                       & 54.06                    & 65.22                       & 44.15                    & 65.22                       & 79.65                       & 46.57                    & 61.76                       & 38.95                    & 61.76                       & 82.44                       \\
Swedish LinES                     & 74.06                    & 74.06                       & 60.21                    & 74.06                       & 84.08                       & 50.16                    & 58.62                       & 40.10                    & 58.62                       & 66.58                       & 55.58                    & 66.39                       & 44.80                    & 66.39                       & 77.01                       \\
Swedish Talbanken                 & 77.72                    & 77.72                       & 62.70                    & 77.91                       & 88.63                       & 58.49                    & 69.06                       & 46.89                    & 69.22                       & 79.32                       & 59.64                    & 69.89                       & 48.41                    & 70.01                       & 81.44                      \\\hline

Arabic PADT & 65.06 & 65.06 & N/A & 66.41 & 77.06             & 51.79 & 53.81 & N/A & 55.01 & 68.54               & 2.89 & 56.34 & N/A & 57.60 & 70.06 \\
Korean GSD & 61.40 & 61.40 & N/A & 61.40 & 85.14                & 47.73 & 54.10 & N/A & 54.10 & 80.75     & 0.30 & 50.50 & N/A & 50.50 & 76.31 \\\hline
\end{tabular}

\caption{Performance of our model compared to UDPipe 1.2 baseline and the winner models of \udst{}.}
\label{table:las}
\end{table*}

\begin{table*}[t]
\setlength{\tabcolsep}{0.3em} 
\scriptsize

\begin{tabular}{|l|c|c|c|c|c|c|c|c|c|c|c|c|c|c|c|}
\hline
\multicolumn{1}{|r|}{Metric}      & \multicolumn{5}{c|}{POS}                                                                                                                      & \multicolumn{5}{c|}{UFeat}                                                                                                                    & \multicolumn{5}{c|}{Lemma}                                                                                                                    \\
\multicolumn{1}{|r|}{Environment} & \multicolumn{2}{c|}{Local}                             & \multicolumn{3}{c|}{TIRA}                                                            & \multicolumn{2}{c|}{Local}                             & \multicolumn{3}{c|}{TIRA}                                                            & \multicolumn{2}{c|}{Local}                             & \multicolumn{3}{c|}{TIRA}                                                            \\
\multicolumn{1}{|r|}{Model}       & \multicolumn{1}{c|}{Our} & \multicolumn{1}{c|}{UDPipe} & \multicolumn{1}{c|}{Our} & \multicolumn{1}{c|}{UDPipe} & \multicolumn{1}{c|}{Winner} & \multicolumn{1}{c|}{Our} & \multicolumn{1}{c|}{UDPipe} & \multicolumn{1}{c|}{Our} & \multicolumn{1}{c|}{UDPipe} & \multicolumn{1}{c|}{Winner} & \multicolumn{1}{c|}{Our} & \multicolumn{1}{c|}{UDPipe} & \multicolumn{1}{c|}{Our} & \multicolumn{1}{c|}{UDPipe} & \multicolumn{1}{c|}{Winner} \\\hline
English EWT                       & 90.47                    & 93.61                       & 92.96                    & 93.62                       & 95.94                       & 93.95                    & 94.60                       & 93.87                    & 94.60                       & 96.03                       & 91.51                    & 95.92                       & 95.77                    & 95.88                       & 97.23                       \\
English GUM                       & 91.00                    & 93.23                       & 89.94                    & 93.24                       & 96.44                       & 93.70                    & 93.89                       & 91.61                    & 93.90                       & 96.68                       & 89.26                    & 94.36                       & 88.66                    & 94.36                       & 96.18                       \\
English LinES                     & 88.93                    & 94.71                       & 87.99                    & 94.71                       & 97.06                       & 92.81                    & 94.97                       & 91.05                    & 94.97                       & 97.08                       & 86.84                    & 95.84                       & 85.37                    & 95.84                       & 96.56                       \\
French Spoken                     & 92.67                    & 92.94                       & 92.18                    & 92.94                       & 97.17                       & 99.97                    & 100.00                      & 100.00                   & 100.00                      & 100.00                      & 86.28                    & 95.84                       & 95.39                    & 95.84                       & 97.50                       \\
French Sequoia                    & 92.81                    & 95.84                       & 95.11                    & 95.84                       & 98.15                       & 93.77                    & 94.97                       & 94.36                    & 94.97                       & 97.50                       & 79.41                    & 97.03                       & 96.56                    & 97.03                       & 97.99                       \\
Finnish TDT                       & 92.72                    & 94.37                       & 92.35                    & 94.37                       & 97.30                       & 89.10                    & 92.06                       & 88.49                    & 92.06                       & 95.58                       & 62.25                    & 86.49                       & 59.50                    & 86.49                       & 95.32                       \\
Finnish FTB                       & 86.77                    & 92.28                       & 86.44                    & 92.28                       & 96.70                       & 89.40                    & 92.74                       & 88.82                    & 92.74                       & 96.89                       & 73.06                    & 88.70                       & 72.35                    & 88.70                       & 97.02                       \\
Swedish LinES                     & 90.02                    & 93.97                       & 89.74                    & 93.97                       & 97.37                       & 83.65                    & 87.23                       & 82.75                    & 87.23                       & 89.61                       & 82.15                    & 94.58                       & 80.59                    & 94.58                       & 96.90                       \\
Swedish Talbanken                 & 91.30                    & 95.35                       & 91.07                    & 95.36                       & 97.90                       & 89.23                    & 94.34                       & 87.93                    & 94.36                       & 96.82                       & 82.99                    & 95.30                       & 81.71                    & 95.28                       & 97.82                      \\\hline
Arabic PADT 
& 88.50 & 89.35 & N/A & 89.34 & 93.63
& 83.07 & 83.39 & N/A & 83.42 & 90.96
&  7.42 & 87.42 & N/A & 87.41 & 91.61 \\

Korean GSD
& 85.34 & 93.44 & N/A & 93.44 & 96.33
& 99.49 & 99.51 & N/A & 99.51 & 99.70
& 12.87 & 87.03 & N/A & 87.03 & 94.02 \\ \hline
\end{tabular}

\caption{Additional metrics describing the performance of our model, UDPipe 1.2 baseline, and the winner models of \udst{}.}
\label{table:pos}
\end{table*}

\section{Discussion}
\subsection{Input vectors for lemma generation}
The initial versions of the lemma decoder did not get the state of the LSTM below $h^3_i$ and positional embedding $\pi_i^j$ as inputs. The network learned to produce lemmas with some accuracy but with many trivial errors. In particular, after training on \textit{English - EWT} treebank, the network learned to remove \textit{s} from the end of the plural nouns. But it also started to produce <end-of-the-word> symbol even if \textit{s} was in the middle of the word. We believe the reason was that there was almost no information available that would allow the decoder to distinguish between plural suffix and a simple \textit{s} inside the word. One could argue that the initial state of the GRU ($h^3_i$) could contain such information, but it could have been lost in the GRU.

To remedy this we decided to pass $h^3_i$ as an input at every step of the decoder. This idea is known to work well in image caption generation. The earliest usage of this trick we know is in \cite{donahue2015long}. 

Additionally, we have added explicit information about the \textit{position} in the word. Unlike \cite{attention-is-all-you-need}, we encode the number of characters left before the end of the word. This choice might be biased towards languages where the ending of the word is the most critical in lemmatization.

By combining these two ideas we got significant improvement in lemma generation for English. We did not do ablation experiments to determine the effect of each of these additions. 

The additional experiments showed that this architecture of the lemmatizer does not generalize to Arabic and Korean. We will investigate this problem in the future work.

\subsection{Balancing different tasks}
Multitask learning in neural networks is usually complicated because of varying difficulty of individual tasks. The $\lambda$ coefficients in (\ref{formula:loss}) can be used to find optimal balance between the tasks. Our initial experiments with all $\lambda$ coefficients equal to $1$ showed that the loss term for POS tagging ($L_p$) had much higher values than the rest. We decided to set $\lambda_p = 0.2$ to give more weight to the other tasks and noticed some improvements in lemma generation.

We believe that more extensive search for better coefficients might help to significantly improve the overall performance of the system.

\subsection{Fighting against overfitting}
The main challenge in training these networks is to overcome overfitting. The only trick we used  was to apply dropout layers before feature extractor LSTMs. We did not apply recurrent dropout \cite{recurrentdropout} or other noise injection techniques, although recent work in language modeling demonstrated the importance of such tricks for obtaining high performance models \cite{awd-lstm}.

\section{Conclusion}
In this paper we have described our submission to \udst{}. Our neural network was learned to jointly produce lemmas, part-of-speech tags and morphological features. It is the first step towards a fully multitask neural architecture that will also produce dependency relations. Future work will include more extensive hyperparameter tuning and experiments with more languages.

\section*{Acknowledgments}
We would like to thank Tigran Galstyan for helpful discussions on neural architecture. We would also like to thank anonymous reviewers for their comments. Additionally, we would like to thank Marat Yavrumyan and Anna Danielyan. Their 
efforts on bringing Armenian into UD family motivated us to work on sentence parsing.


\bibliography{udst2018}

\begin{thebibliography}{}
\expandafter\ifx\csname natexlab\endcsname\relax\def\natexlab#1{#1}\fi

\bibitem[{Bojanowski et~al.(2017)Bojanowski, Grave, Joulin, and
  Mikolov}]{fasttext}
Piotr Bojanowski, Edouard Grave, Armand Joulin, and Tomas Mikolov. 2017.
\newblock Enriching word vectors with subword information.
\newblock {\em Transactions of the Association for Computational Linguistics\/}
  5:135--146.

\bibitem[{Chen and Manning(2014)}]{ChenManning}
Danqi Chen and Christopher Manning. 2014.
\newblock A fast and accurate dependency parser using neural networks.
\newblock In {\em Proceedings of the 2014 conference on empirical methods in
  natural language processing (EMNLP)\/}. pages 740--750.

\bibitem[{Chiu and Nichols(2016)}]{NERBiLSTMCNN}
Jason~P.C. Chiu and Eric Nichols. 2016.
\newblock \href{https://transacl.org/ojs/index.php/tacl/article/view/792}{Named
  entity recognition with bidirectional lstm-cnns}.
\newblock {\em Transactions of the Association for Computational Linguistics\/}
  4:357--370.
\newblock
  \href{https://transacl.org/ojs/index.php/tacl/article/view/792}{https://transacl.org/ojs/index.php/tacl/article/view/792}.

\bibitem[{Cho et~al.(2014)Cho, van Merrienboer, Gulcehre, Bahdanau, Bougares,
  Schwenk, and Bengio}]{gru}
Kyunghyun Cho, Bart van Merrienboer, Caglar Gulcehre, Dzmitry Bahdanau, Fethi
  Bougares, Holger Schwenk, and Yoshua Bengio. 2014.
\newblock \href{https://doi.org/10.3115/v1/D14-1179}{Learning phrase
  representations using rnn encoder--decoder for statistical machine
  translation}.
\newblock In {\em Proceedings of the 2014 Conference on Empirical Methods in
  Natural Language Processing (EMNLP)\/}. Association for Computational
  Linguistics, pages 1724--1734.
\newblock
  \href{https://doi.org/10.3115/v1/D14-1179}{https://doi.org/10.3115/v1/D14-1179}.

\bibitem[{Donahue et~al.(2015)Donahue, Anne~Hendricks, Guadarrama, Rohrbach,
  Venugopalan, Saenko, and Darrell}]{donahue2015long}
Jeffrey Donahue, Lisa Anne~Hendricks, Sergio Guadarrama, Marcus Rohrbach,
  Subhashini Venugopalan, Kate Saenko, and Trevor Darrell. 2015.
\newblock Long-term recurrent convolutional networks for visual recognition and
  description.
\newblock In {\em Proceedings of the IEEE conference on computer vision and
  pattern recognition\/}. pages 2625--2634.

\bibitem[{Dozat et~al.(2017)Dozat, Qi, and Manning}]{StanfordCoNLL2017}
Timothy Dozat, Peng Qi, and Christopher~D. Manning. 2017.
\newblock \href{https://doi.org/10.18653/v1/K17-3002}{Stanford's graph-based
  neural dependency parser at the conll 2017 shared task}.
\newblock In {\em Proceedings of the CoNLL 2017 Shared Task: Multilingual
  Parsing from Raw Text to Universal Dependencies\/}. Association for
  Computational Linguistics, pages 20--30.
\newblock
  \href{https://doi.org/10.18653/v1/K17-3002}{https://doi.org/10.18653/v1/K17-3002}.

\bibitem[{Gal and Ghahramani(2016)}]{recurrentdropout}
Yarin Gal and Zoubin Ghahramani. 2016.
\newblock A theoretically grounded application of dropout in recurrent neural
  networks.
\newblock In {\em Advances in neural information processing systems\/}. pages
  1019--1027.

\bibitem[{Gehring et~al.(2017)Gehring, Auli, Grangier, Yarats, and
  Dauphin}]{convs2s:fairseq2seq}
Jonas Gehring, Michael Auli, David Grangier, Denis Yarats, and Yann~N Dauphin.
  2017.
\newblock {Convolutional Sequence to Sequence Learning}.
\newblock {\em ArXiv e-prints\/} .

\bibitem[{Hochreiter and Schmidhuber(1997)}]{lstm}
Sepp Hochreiter and J{\"u}rgen Schmidhuber. 1997.
\newblock Long short-term memory.
\newblock {\em Neural computation\/} 9(8):1735--1780.

\bibitem[{Lample et~al.(2016)Lample, Ballesteros, Kawakami, Subramanian, and
  Dyer}]{LSTMCRFNER}
Guillaume Lample, Miguel Ballesteros, Kazuya Kawakami, Sandeep Subramanian, and
  Chris Dyer. 2016.
\newblock Neural architectures for named entity recognition.
\newblock In {\em Proc. NAACL-HLT\/}.

\bibitem[{Merity et~al.(2018)Merity, Keskar, and Socher}]{awd-lstm}
Stephen Merity, Nitish~Shirish Keskar, and Richard Socher. 2018.
\newblock \href{https://openreview.net/forum?id=SyyGPP0TZ}{Regularizing and
  optimizing {LSTM} language models}.
\newblock In {\em International Conference on Learning Representations\/}.
\newblock
  \href{https://openreview.net/forum?id=SyyGPP0TZ}{https://openreview.net/forum?id=SyyGPP0TZ}.

\bibitem[{Nguyen et~al.(2017)Nguyen, Dras, and Johnson}]{jPTDP1}
Dat~Quoc Nguyen, Mark Dras, and Mark Johnson. 2017.
\newblock \href{http://www.aclweb.org/anthology/K17-3014}{A novel neural
  network model for joint pos tagging and graph-based dependency parsing}.
\newblock In {\em Proceedings of the CoNLL 2017 Shared Task: Multilingual
  Parsing from Raw Text to Universal Dependencies\/}. Association for
  Computational Linguistics, Vancouver, Canada, pages 134--142.
\newblock
  \href{http://www.aclweb.org/anthology/K17-3014}{http://www.aclweb.org/anthology/K17-3014}.

\bibitem[{Nivre et~al.(2016)Nivre, de~Marneffe, Ginter, Goldberg, Haji{\v{c}},
  Manning, McDonald, Petrov, Pyysalo, Silveira, Tsarfaty, and Zeman}]{ud}
Joakim Nivre, Marie-Catherine de~Marneffe, Filip Ginter, Yoav Goldberg, Jan
  Haji{\v{c}}, Christopher Manning, Ryan McDonald, Slav Petrov, Sampo Pyysalo,
  Natalia Silveira, Reut Tsarfaty, and Daniel Zeman. 2016.
\newblock {Universal Dependencies} v1: A multilingual treebank collection.
\newblock In {\em Proceedings of the 10th International Conference on Language
  Resources and Evaluation ({LREC} 2016)\/}. European Language Resources
  Association, Portorož, Slovenia, pages 1659--1666.

\bibitem[{Nivre et~al.(2018)}]{ud22data}
Joakim Nivre et~al. 2018.
\newblock \href{http://hdl.handle.net/11234/1-1983xxx}{{Universal Dependencies
  2.2}}.
\newblock {LINDAT}/{CLARIN} digital library at the Institute of Formal and
  Applied Linguistics, Charles University, Prague,
  \url{http://hdl.handle.net/11234/1-1983xxx}.
\newblock
  \href{http://hdl.handle.net/11234/1-1983xxx}{http://hdl.handle.net/11234/1-1983xxx}.

\bibitem[{Potthast et~al.(2014)Potthast, Gollub, Rangel, Rosso, Stamatatos, and
  Stein}]{tira}
Martin Potthast, Tim Gollub, Francisco Rangel, Paolo Rosso, Efstathios
  Stamatatos, and Benno Stein. 2014.
\newblock \href{https://doi.org/10.1007/978-3-319-11382-1\_{}22}{Improving the
  reproducibility of {PAN}'s shared tasks: Plagiarism detection, author
  identification, and author profiling}.
\newblock In Evangelos Kanoulas, Mihai Lupu, Paul Clough, Mark Sanderson, Mark
  Hall, Allan Hanbury, and Elaine Toms, editors, {\em Information Access
  Evaluation meets Multilinguality, Multimodality, and Visualization. 5th
  International Conference of the {CLEF} Initiative ({CLEF} 14)\/}. Springer,
  Berlin Heidelberg New York, pages 268--299.
\newblock
  \href{https://doi.org/10.1007/978-3-319-11382-1\_{}22}{https://doi.org/10.1007/978-3-319-11382-1\_{}22}.

\bibitem[{Reimers and Gurevych(2017)}]{Reimers:2017:EMNLP}
Nils Reimers and Iryna Gurevych. 2017.
\newblock \href{http://aclweb.org/anthology/D17-1035}{{Reporting Score
  Distributions Makes a Difference: Performance Study of LSTM-networks for
  Sequence Tagging}}.
\newblock In {\em Proceedings of the 2017 Conference on Empirical Methods in
  Natural Language Processing (EMNLP)\/}. Copenhagen, Denmark, pages 338--348.
\newblock
  \href{http://aclweb.org/anthology/D17-1035}{http://aclweb.org/anthology/D17-1035}.

\bibitem[{Straka et~al.(2016)Straka, Haji\v{c}, and Strakov\'{a}}]{udpipe}
Milan Straka, Jan Haji\v{c}, and Jana Strakov\'{a}. 2016.
\newblock {UDPipe:} trainable pipeline for processing {CoNLL-U} files
  performing tokenization, morphological analysis, {POS} tagging and parsing.
\newblock In {\em Proceedings of the 10th International Conference on Language
  Resources and Evaluation ({LREC} 2016)\/}. European Language Resources
  Association, Portorož, Slovenia.

\bibitem[{Strubell et~al.(2017)Strubell, Verga, Belanger, and
  McCallum}]{ERDilatedConv}
Emma Strubell, Patrick Verga, David Belanger, and Andrew McCallum. 2017.
\newblock Fast and accurate entity recognition with iterated dilated
  convolutions.
\newblock In {\em Proceedings of the 2017 Conference on Empirical Methods in
  Natural Language Processing\/}. pages 2670--2680.

\bibitem[{Sukhbaatar et~al.(2015)Sukhbaatar, Weston, Fergus
  et~al.}]{end2end-memory-networks}
Sainbayar Sukhbaatar, Jason Weston, Rob Fergus, et~al. 2015.
\newblock End-to-end memory networks.
\newblock In {\em Advances in neural information processing systems\/}. pages
  2440--2448.

\bibitem[{Vaswani et~al.(2017)Vaswani, Shazeer, Parmar, Uszkoreit, Jones,
  Gomez, Kaiser, and Polosukhin}]{attention-is-all-you-need}
Ashish Vaswani, Noam Shazeer, Niki Parmar, Jakob Uszkoreit, Llion Jones,
  Aidan~N Gomez, {\L}ukasz Kaiser, and Illia Polosukhin. 2017.
\newblock Attention is all you need.
\newblock In {\em Advances in Neural Information Processing Systems\/}. pages
  5998--6008.

\bibitem[{Zeman et~al.(2018{\natexlab{a}})}]{ud22testdata}
Dan Zeman et~al. 2018{\natexlab{a}}.
\newblock \href{http://hdl.handle.net/11234/1-2184}{{Universal Dependencies 2.2
  – CoNLL} 2018 shared task development and test data}.
\newblock {LINDAT}/{CLARIN} digital library at the Institute of Formal and
  Applied Linguistics, Charles University, Prague,
  \url{http://hdl.handle.net/11234/1-2184}.
\newblock
  \href{http://hdl.handle.net/11234/1-2184}{http://hdl.handle.net/11234/1-2184}.

\bibitem[{Zeman et~al.(2018{\natexlab{b}})Zeman, Haji{\v{c}}, Popel, Potthast,
  Straka, Ginter, Nivre, and Petrov}]{udst:overview}
Daniel Zeman, Jan Haji{\v{c}}, Martin Popel, Martin Potthast, Milan Straka,
  Filip Ginter, Joakim Nivre, and Slav Petrov. 2018{\natexlab{b}}.
\newblock {CoNLL 2018 Shared Task: Multilingual Parsing from Raw Text to
  Universal Dependencies}.
\newblock In {\em Proceedings of the CoNLL 2018 Shared Task: Multilingual
  Parsing from Raw Text to Universal Dependencies\/}. Association for
  Computational Linguistics, Brussels, Belgium, pages 1--20.

\end{thebibliography}

\bibliographystyle{acl_natbib}







\end{document}